%% file: Example.tex
\documentclass[a4paper,twoside]{article}

\usepackage{epsfig}
\usepackage{subcaption}
\usepackage{calc}
\usepackage{amssymb}
\usepackage{amstext}
\usepackage{amsmath}
\usepackage{amsthm}
\usepackage{multicol}
\usepackage{mathptmx}
\usepackage{apalike}
\usepackage{algorithm2e}
\usepackage[bottom]{footmisc}

\usepackage{url}
\usepackage{lineno}
\usepackage{tabularx}
\usepackage{xcolor}
\usepackage{color}

\usepackage{adjustbox}

\usepackage{tikz, pgfplots}
\usetikzlibrary{calc,pgfplots.groupplots,arrows.meta,positioning,fillbetween,shapes.geometric,shapes,plotmarks,fit,backgrounds,trees,patterns}

\pgfmathdeclarefunction{elu}{6}{\pgfmathparse{
		#1 > #3 ? #2 * (#1 - #3)+#4 :
		(#6 * (#1 - (ln(#6/#2)+#3+#5*#2/#6-1))+#4)
}}

\usepgfplotslibrary{external} 

\newcommand{\tightcolorbox}[2]{%
	\begingroup
	\setlength{\fboxsep}{1pt}
	\colorbox{#1}{\strut #2}%
	\endgroup
}

\pgfplotsset{
	title style = {font=\normalsize},
	every tick/.style={
		black,
	},
	y tick label style={/pgf/number format/fixed,xshift=.2em},
	y label style={yshift=-.5em},
	x tick label style={/pgf/number format/fixed,yshift=.2em},
	x label style={yshift=.5em},
	label style={font=\footnotesize},
	every tick label/.append style={font=\scriptsize},
	xtick pos=left,
	ytick pos=left,
	tick align = center,
	legend style={inner xsep=2pt, inner ysep=1pt, draw=black,font=\footnotesize},
	legend columns=-1,
	/pgfplots/ybar legend/.style={
		/pgfplots/legend image code/.code={%
			\draw[##1,/tikz/.cd,bar width=3pt,yshift=-0.2em,bar shift=0pt]
			plot coordinates {(0cm,0.7em)};},},
	ownBar/.style={%
		legend image code/.code={
			\draw[fill=##1,line width=0.5pt]
			(0cm,-0.05cm) rectangle ++(0.1cm,0.2cm);}},
}

\definecolor{c0}{rgb}{0.20660,0.47430,0.99260}
\definecolor{c1}{rgb}{0.678,0.847,0.902} 
\definecolor{c2}{rgb}{0.5,0.5,0}
\definecolor{c3}{rgb}{1,0.625,0}
\definecolor{c4}{rgb}{1,0,0.75}
\definecolor{c5}{rgb}{0.180,0.545,0.341} 
\definecolor{c6}{rgb}{0.827,0.827,0.827} 
\definecolor{c7}{rgb}{0,0.875,1}
\definecolor{c8}{rgb}{1,0,0}
\definecolor{c9}{RGB}{50,51,152}
\definecolor{c10}{rgb}{63,149,38}
\definecolor{c11}{RGB}{214,0,45}
\definecolor{c12}{RGB}{255,152,0}
\definecolor{c13}{RGB}{148,32,146}
\definecolor{c14}{RGB}{192,192,192}

\definecolor{c201}{rgb}{0.679,0.847,0.902} 
\definecolor{c202}{rgb}{0.547,0.725,0.801}
\definecolor{c203}{rgb}{0.416,0.603,0.700}
\definecolor{c204}{rgb}{0.285,0.480,0.600}
\definecolor{c205}{rgb}{0.154,0.358,0.500}
\definecolor{c206}{rgb}{0.023,0.235,0.400}
\definecolor{c207}{rgb}{-0.108,0.113,0.300}
\definecolor{c208}{rgb}{-0.239,-0.014,0.200}

\definecolor{c209}{rgb}{0.01176,0.01176,0.52157}
\definecolor{c210}{rgb}{0.97255,0.69020,0.00000}
\definecolor{c211}{rgb}{0.20660,0.47430,0.99260}
\definecolor{c212}{rgb}{0.06380,0.74460,0.72920}

\tikzset{
	block/.style = {draw, rectangle, rounded corners, thick, minimum height=2em, minimum width=3em},
	output/.style = {coordinate},
	input/.style = {coordinate}, 
	sum/.style = {draw, circle, minimum size=0.5cm, inner sep=0pt} 
}

\usepackage{graphicx}      
\usepackage{natbib}        
\usepackage{SCITEPRESS}     
\begin{document}

\title{Space-Filling Regularization for Robust and Interpretable\\Nonlinear State Space Models} 


\author{\authorname{Hermann Klein\orcidAuthor{0009-0009-5386-3575}, Max Heinz Herkersdorf\orcidAuthor{0009-0007-8554-8153} and Oliver Nelles\orcidAuthor{0000-0002-9471-8106}}
	\affiliation{University of Siegen, Department Mechanical Engineering, Automatic Control – Mechatronics, Paul-Bonatz-Str.\ 9-11, 57068 Siegen, Germany}
	\email{\{hermann.klein, max.herkersdorf, oliver.nelles\}@uni-siegen.de}
}


\abstract{The state space dynamics representation is the most general approach for nonlinear systems and often chosen for system identification. During training, the state trajectory can deform significantly leading to poor data coverage of the state space. This can cause significant issues for space-oriented training algorithms which e.g. rely on grid structures, tree partitioning, or similar. Besides hindering training, significant state trajectory deformations also deteriorate interpretability and robustness properties. This paper proposes a new type of space-filling regularization that ensures a favorable data distribution in state space via introducing a data-distribution-based penalty. This method is demonstrated in local model network architectures where good interpretability is a major concern. The proposed approach integrates ideas from modeling and design of experiments for state space structures.
	This is why we present two regularization techniques for the data point distributions of the state trajectories for local affine state space models.
	Beyond that, we demonstrate the results on a widely known system identification benchmark.}

\onecolumn \maketitle \normalsize \setcounter{footnote}{0} \vfill

\section{\uppercase{Introduction}}
\label{sec:Introduction}
Models of dynamic systems are the base for numerous industrial applications, especially for control design or virtual sensing.
Insofar as white box models derived from physical principles
oftentimes fail to meet an acceptable tradeoff between modeling effort and accuracy, data-driven model learning, also known as system identification, comes into play. Here, the user can choose between different mathematical architectures to describe the system's dynamics. A state space representation is the most abstract and general kind as well as required for many control design methods. Examples of alternative approaches are feedforward-trained models like finite impulse response or nonlinear autoregressive models with exogenous input or output feedback-trained kinds like output error models~\citep{Ljung1999}. In comparison to the latter and especially for the nonlinear case, the state space model is a more powerful modeling approach due to its internal state feedback~\citep{Belz2017,Schuessler2019a}.

Whereas \textit{linear} state space models can be extracted from frequency domain approaches with subspace identification~\citep{Pintelon2012,VanOverschee1995}, \textit{nonlinear} models have to be trained via nonlinear optimization with the objective of simulation error fit. They generally arise when the linear state and output functions are replaced with nonlinear function approximators. Here, various approaches exist, including polynomials~\citep{Paduart2010} or neural networks~\citep{Suykens1995,Forgione2021}. Due to local interpretability and explainability reasons, we focus on the class of local affine state space models. Mathematically related architectures are (softly switching) piecewise affine~\citep{Garulli2012}, Takagi Sugeno (TS) or quasi-Linear Parameter Varying (qLPV) state space models~\citep{Rotondo2015}.

In general, system identification tries to find an accurate model based on input/output training datasets~\citep{Ljung2020}. As a consequence, an immanent feature of nonlinear state space models is the implicit optimization of the model's state, since it depends on trainable parameters. Due to the simulation error fit, the state is optimized indirectly. In geometrical terms, deformations of the trajectories can be observed within the optimization procedure. This can be visualized in the phase plot of the state variables' trajectories, at least for the two-dimensional projection\footnote{For simplicity, the phase plot of the state variables is designated as the \textit{state trajectory} in this paper.}. The point distribution in the state space is manipulated until the termination of the optimization. For TS models, 
the effects on the state trajectory can be connected with local regions. This interaction can affect robustness and interpretability and is addressed in this contribution.

As a first try, any system identification task can be challenged with a pure data-driven (black box) approach. In such a setting, the state space model is fully parametrized including redundant parameters. A common technique in estimation theory is to use regularization by application of an additional penalty term to the objective function~\citep{Boyd2004}. In the context of nonlinear state space modeling, recent work deals with L1 regularization for black box models~\citep{bemporad2024lbfgsbapproachlinearnonlinear}. Moreover,~\citet{liu2024physicsguidedstatespacemodelaugmentation} use regularization to make a black-box state meet a physical-interpretable state realization.
An overview on methods for nonlinear systems can e.g. be found in~\citet{Pillonetto2022}.

As mentioned above, the interaction of the data point distribution with local activation functions is crucial for models with regional organization, like the Local Model State Space Network (LMSSN). As a consequence, we develop a \textit{space-filling} regularization for the LMSSN procedure. This transfers ideas from the Design of Experiments into model training. The novelty of this contribution is the development of a space-filling penalty term. We analyze it in detail and compare two related regularization techniques.

The article is organized as follows. In Section~\ref{sec:NonlinearStateSpace}, we present the LMSSN-based modeling procedure. Next, we present the evolution of the space-filling indicators within the optimization procedure in Section~\ref{sec:SpaceFilling}. The effect of a space-filling penalty term is presented in Section~\ref{sec:Regularization}. Our method is tested on a benchmark process in Section~\ref{sec:CaseStudy}. Finally, we summarize our work and give an outlook on future lines of research.

\section{\uppercase{Nonlinear State Space Modeling}}\label{sec:NonlinearStateSpace}
The proposed method is suitable for all state space models in which the state explores an a priori known range. 
Since we focus on the LMSSN method, its functionality is introduced in this section.
This unregularized version serves as a reference for recent advancements regarding space-filling regularization.

\subsection{Local Model State Space Network}\label{sec:LMSSN}
LMSSN is a system identification method to create discrete-time nonlinear state space models. The mathematical model architecture embeds Local Model Networks (LMNs) in a state space framework. LMSSN modeling results in local affine state space models, weighted with normalized radial basis functions (NRBF). Detailed information can be found in~\citet{Schuessler2022}. It is emphasized that LMSSN differs from TS and qLPV approaches in the use of the Local Linear Model Tree (LOLIMOT). The tree-construction algorithm serves for the determination of the fuzzy premise variables (or scheduling variables in qLPV terms). Additionally, it uses a heuristic axis-orthogonal input space partitioning approach by automatic parametrization of the validity functions. For more information about LOLIMOT, we refer to~\citet{Nelles2020}.

The resulting LMSSN model is formulated with the following equations,
\begin{equation}\label{eq:LMSSNt_NN}
	\begin{split}
		\underline{\hat x}(k+1) &= \sum_{j=1}^{n_{\text{LM},x}} (\underline{A}_{j} \underline{{\hat{x}}}(k) + \underline{b}_{j} {u}(k) + \underline{o}_{j})\circ {{\Phi}}^{[x]}_{j}(k)\\
		{{\hat y}}(k) &= \sum_{j=1}^{n_{\text{LM},y}} (\underline{c}_{j}^\top \underline{{\hat{x}}}(k) + {d}_{j} {u}(k) + {p}_{j})\circ {{\Phi}}^{[y]}_{j}(k).
	\end{split}
\end{equation}
For the single-input single-output case ($n_u$=1 input and $n_y$=1 output variable), the tensors $\underline{A}_{j} \in \mathbb{R}^{n_x \times n_x}$, $\underline{b}_{j} \in \mathbb{R}^{n_x}$, $\underline{o}_{j} \in \mathbb{R}^{n_x}$, $\underline{c}_{j} \in \mathbb{R}^{n_x}$, ${d}_{j} \in \mathbb{R}$ and ${p}_{j} \in \mathbb{R}$ are filled with the slopes and offsets of the $j$-th local affine model (LM). The dynamical order is expressed by the number of state variables $n_x$ and
the discrete-time is indicated with~$k$.
In configuration~\eqref{eq:LMSSNt_NN}, there are altogether $n_{\text{LM},x}$ superposed affine models for the state~$\underline{\hat x}(k+1)$ and $n_{\text{LM},y}$ for the output~$\hat{y}(k)$.
The basis functions $\Phi^{[x]}_{j}$ and $\Phi^{[y]}_{j}$ express the $j$-th local validity function. Since they are implemented with normalized Gaussians, the partition of unity holds, $\sum_{j=1}^{n_{\text{LM},x}}\Phi^{[x]}_{j}=1$ and $\sum_{j=1}^{n_{\text{LM},y}}\Phi^{[y]}_{j}=1$.

Due to the design of LMSSN in state space form, the parameter initialization can be done deterministically with a global linear state space model. The latter is gathered as the Best Linear Approximation (BLA) of the process in the frequency domain~\citep{McKelvey1996}. Next, subspace-based identification converts the \textit{nonparametric} frequency response function model into a \textit{parametric} linear state space model. The fully-parametrized discrete-time model is then transformed into a balanced realization via a similarity transformation~\citep{Verriest1983, Luenberger1967}. It evokes a diagonal structure within the state space matrices which leads to a state realization with an evenly (and thus desired) trajectory shape.

After the linear model initialization, the LOLIMOT algorithm starts. With the help of the stepwise addition of weighted LMs, it enables for nonlinear curve approximation. Each LM is connected with a region of activation in the extended input/state space~$\underline{\tilde{u}}=[\underline{\hat{x}},{u}]^{\text{T}}$. This corresponds to the partition of $\tilde{\underline{u}}$ with interpolation regimes for smooth cross-fading in between. It can be seen as a generation of \textit{splits} of $\underline{\tilde{u}}$, arranged by center placement of the NRBFs. If every dimension of $\tilde{\underline{u}}$ is to be analyzed in form of a split, $n_\text{Split}$=dim($\tilde{\underline{u}}$) split options arise. For the ease of the center placement, it is intuitive to scale the state trajectory into a fixed interval, preferably the unit cube. The scaling is actualized with an affine scaling transformation, parametrized with range and offset values according to minimum and maximum of $\hat{\underline{x}}$. The procedure is called \textit{split-adaption algorithm} according to~\citet{Schuessler2019a}. Each LOLIMOT iteration investigates all $n_\text{Split}$ options for the region with the largest error separately via an optimization run.\footnote{Note the LOLIMOT iterations have to be distinguished from the gradient-based iterations within the nonlinear optimization of each split option.}


\subsection{Problem Statement}\label{sec:Trajectory}
The loss function $J$ being optimized in each LOLIMOT iteration is the sum of the squared output errors. For a one-dimensional output variable, the optimization problem is stated as
\begin{equation} \label{eq:nlp}
	\begin{split}
		\underset{\underline{\theta}}{\text{min}}\ &J(\underline{\theta})=
		\underset{\underline{\theta}}{\text{min}}  \sum_{k = 1}^{N} \Bigl({y}(k )-{\hat{y}}(k;\underline{\theta})\Bigr)^2\\
		\text{s.t.}
		&\quad{\underline{{\hat{x}}}}(k+1) = \text{LMN}(\underline{\hat{x}}(k),{u}(k);{\underline{\theta}}^{[x]})\\
		&\quad{{\hat{y}}}(k) = \text{LMN}({\underline{{\hat{x}}}}(k),{u}(k);{\underline{\theta}}^{[y]}),
	\end{split}
\end{equation}
where $\underline \theta$ is the vector of all model parameters and $N$ is the number of samples collected for training. The nonlinear problem is solved with a Quasi-Newton optimizer with BFGS Hessian approximation. The optimization is run till convergence in a local minimum terminating on a loss and gradient tolerance or minimal step size.

\begin{figure}[htb!]
	\begin{center}
		\scalebox{0.6}{\input{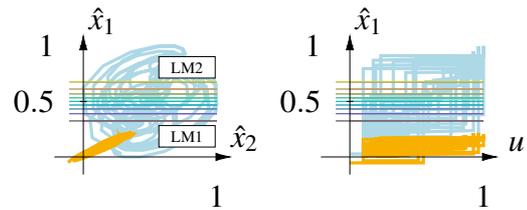}}
	\end{center}
	\caption{Extended state trajectory before~(\tightcolorbox{c1}{ }) and after~(\tightcolorbox{c210}{ }) the optimization of the first split in the~$\hat{x}_1$-dimension. The colored contour lines show the validity functions and mark the interpolation regime. For brevity, the $\hat{x}_2$-$u$ projection is left out.}
	\label{fig:xTraj}
\end{figure}

By fitting the parameters according to the minimization of the output error, the state trajectory changes its shape from the balanced realization to the one that is most effective for the objective~\eqref{eq:nlp}. Next, we show a possible consequence on the state trajectory for the first split of a simulated second-order single-input single-output system with nonlinear feedback,
\begin{equation} \label{eq:process}
	\ddot{y}(t) + a_1 \dot{y}(t) + a_0 f({y}(t)) = b_0 u(t),
\end{equation}
with the parameter values $a_0$=$b_0$=15 and $a_1$=3. The nonlinear feedback $f(y)$ is arranged with a smooth exponentially rising curve. Figure~\ref{fig:xTraj} demonstrates that the initial state trajectory was within the unit cube, whereas the optimized one is compressed. Obviously, the nonlinear behavior of the process can be captured although no data point is fully active in the upper LM. Nevertheless, LM2 influences the data points via smooth cross-fading of both LMs in the interpolation regime. The shown behavior causes problems regarding interpretation. Furthermore, the effect is accompanied by robustness problems. Large local parameter values of LM2 are necessary to influence data, which are located in the lower LM1. With the help of the maximum absolute value of the eigenvalues~$\mu_i$ of the local state transition matrices,
\begin{equation*}
	\underline{A}_{\text{LM,}j}=
	\begin{bmatrix}
		\underline{a}_{1,j}^\top\\
		\underline{a}_{2,j}^\top
	\end{bmatrix},
\end{equation*}
unstable poles in LM2 can be detected:
\begin{equation*} \label{eq:poles_unreg}
	\begin{split}
		&\underset{i}{\max} \left| \mu_i(\underline{A}_{\text{LM,}1}) \right| = 0.99<1,\\
		&\underset{i}{\max} \left| \mu_i(\underline{A}_{\text{LM,}2}) \right| =4.23>1
	\end{split}
\end{equation*}
The phenomenon can be seen as overfitting and contains the risk of unstable extrapolation behavior. Finally, the risk of reactivation~\citep{Nelles2020} of LMs within the next splits is increased, since the split adaption leads to major differences in the NRBF's standard deviations.

Note that the fixed centers of the LOLIMOT-partitioned state space make this effect visible. It can hardly be investigated for deep neural state space models or recurrent neural networks, where all neural weights are trainable parameters. Furthermore, it must be mentioned that the observed behavior (Fig.~\ref{fig:xTraj}) enables modeling with low flexibility. In the case of LMSSN this means that only one split is sufficient for the complete identification. Thus it is very effective because only two local \textit{linear} models are able to fit a process with a smooth curve~\eqref{eq:process}.


\section{\uppercase{Space-Filling Evolution}}\label{sec:SpaceFilling}
Before merging modeling with space-filling properties, three suitable metrics for the assessment of data point distributions are presented. Afterwards, their evolution along the optimization progress is demonstrated. For mathematical reasons, the state trajectory is described as the set of points in the extended input/state space as
\begin{equation*}
	\mathcal{X}=\underline{\tilde{u}}(k), \forall k \in \{1, \dots, N\}.
\end{equation*}

\subsection{Metrics and Indicators}\label{sec:Metrics}
Inspired by the shape of the state trajectory in Fig.~\ref{fig:xTraj}, the covered volume is an intuitive and straightforward space-filling indicator. It can be assessed as the volume of a convex hull on the state trajectory (CHV)~\citep{Boyd2004}. Written for the extended input/state space, the volume~$V$ is defined as
\begin{equation}
	\begin{split}
		V = \frac{1}{\text{dim}(\underline{\tilde{u}})}\sqrt{\text{det}(\underline{G})},
	\end{split}
\end{equation}
where the Gram matrix~$\underline{G}$ is filled with
\begin{equation*}
	G_{i,j} = \underline{\tilde{u}}(i)^{\mathrm{T}} \, \underline{\tilde{u}}(j),\ \forall i, j \in \mathcal{H}.
\end{equation*}
The convex hull is defined by the set of points~$\mathcal{H} \subseteq \{1, 2, \dots, N\}$ derived with the help of the quickhull algorithm. Intuitively, the CHV is going to deliver large values for widely distributed points and small values for concentrated points.

Alternatively, support point-based indicators are commonly applied for assessing space-filling measurements. Here, the state trajectory is compared to a given distribution of~$n_g$ grid or Sobol points $\mathcal{G}$. In this contribution, we concentrate on a uniform grid, implying that a uniform distribution is desired. This leads to the final two metrics.

First, the mean of the minimum distances~$\psi_p$ from every grid point~$\underline{g}_j , \forall j \in \{j = 1,2,\dots,n_g\}$, to the nearest extended input/state space point~$\underline{\tilde{u}}(k)$ is calculated~\citep{Herkersdorf2025}. The criterion is strongly related to Monte Carlo Uniform Sampling Distribution Approximation~\citep{Smits2024}. In this paper, it is denoted with $\psi_p$ and given as
\begin{equation}\label{eq:psip}
	\psi_p = \frac{1}{N} \sum_{j=1}^{M} \min_{k \in \{1, \dots, N\}} d(\underline{\tilde{u}}(k), \underline{g}_j)
\end{equation}
with the Euclidian distance $d$
\begin{equation*}
	d(\underline{\tilde{u}}(k), \underline{g}_j) = \sqrt{\underline{\tilde{u}}(k) - \underline{g}_j)^{\text{T}}(\underline{\tilde{u}}(k) - \underline{g}_j)}.
\end{equation*}

Secondly, from a statistical point of view, the Kullback-Leibler Divergence (KLD) is able to assess the similarity of two probability density functions (PDF)~\cite{Kullback1951} The PDF of a dataset~$\mathcal{S}$ is denoted by~$P(\mathcal{S})$. In the prevalent application, it compares the estimated probability density~$\hat{P}(\mathcal{X})$ with the uniform distribution of a grid~$P(\mathcal{G})=1/n_g$. The KLD is then calculated as
\begin{equation}
	{\mathrm{KLD}} = -\frac{1}{n_g}\sum_{k=1}^{N} \log \left(\hat{P}(\mathcal{X})\right).
\end{equation}
The required probability density estimation of the state trajectory is done with kernel density estimation~\citep{Scott1992}. 
For the measurement of the space-filling properties of a point distribution, both KLD and~$\psi_{p}$ will decrease when the state trajectory becomes uniformly, i.e., meets the grid.


\subsection{Tracking within Nonlinear Optimization}\label{sec:Tracking}
\begin{figure}[htb!]
	\begin{center}
		\scalebox{0.5}{\input{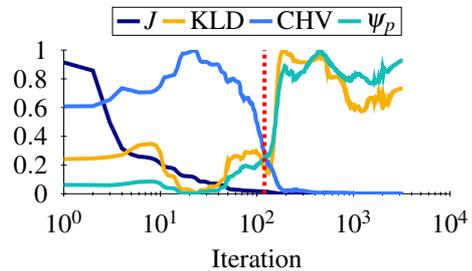}}
	\end{center}
	\caption{Tracking of loss and space-filling indicators according to the state trajectories depicted in Fig.~\ref{fig:xTraj}.}
	\label{fig:SF_Evo}
\end{figure}

For the sake of plausibility, the previously introduced indicators in the example depicted in Fig.~\ref{fig:xTraj} are examined. Obviously,~CHV decreases as the optimization progresses. On the other hand, KLD and~$\psi_p$ are expected to grow, because the state trajectory is evenly distributed \textit{before} the optimization, and narrow-shaped \textit{after} the optimization has converged. Figure~\ref{fig:SF_Evo} shows the evolution of the three space-filling indicators as well as the loss along with the iterations.

It can be stated that all indicators detect the shrinkage of the state trajectory and thus are validated. Note that the absolute numerical values of the indicators are not of interest here. The focus is only on the change during the optimization. This is why the curves are normalized w.r.t. their maximum values.

On closer inspection, it can be observed that the changing curvature reveals large rates around iteration 110, indicated by the vertical line. Loosely spoken, the optimization has already reached an acceptable fit of the process output at this point which is visible in the loss curve. It can be concluded that space-filling measured by those indices gets worse \textit{after} the model performance met a satisfying level. This confirms that the phenomenon of state trajectory deformation can be seen as an overfitting of nonlinear state space models. This is why we recommend a regularization technique with an early-stopping effect in the next section.


\section{\uppercase{Regularized Identification}}\label{sec:Regularization}
Since strong state trajectory deformation is an unsatisfying effect, it is reasonable to design an additional penalty for the objective that drives its data points to a desired uniform distribution. In this section, we present two penalty term approaches for space-filling regularization of problem~\eqref{eq:nlp}. Both approaches use the~$\psi_{p}$ described by~\eqref{eq:psip} which was chosen due to implementation reasons. Generally, Fig.~\ref{fig:SF_Evo} suggests that all space-filling indicators would lead to similar results.

\subsection{Space-Filling Penalty Term}\label{sec:Penalty}
\begin{figure}[htb!]
	\begin{center}
		\scalebox{0.6}{\input{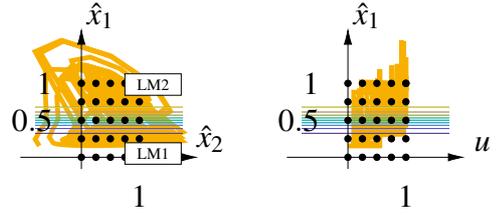}}
	\end{center}
	\caption{State trajectory after the optimization for $\lambda=10$ on $n_g$=125 grid points in three-dimensional extended input/state space.}
	\label{fig:xTraj_reg}
\end{figure}

As a consequence of the~$\psi_{p}$ indicator delivering low numerical values when data are space-filling and high values when data are non-uniformly distributed, it seems natural to add~$\psi_{p}^2$ to the sum of the squared output errors~$J$ in a weighted manner. The resulting regularized problem is
\begin{equation} \label{eq:regopt}
	\begin{split}
		\underset{\underline{\theta}}{\text{min}}\  &J(\underline{\theta})+ \lambda \cdot \psi_p(\underline{\theta})^2\\
		&\text{s.t. Dynamics in \eqref{eq:nlp}},
	\end{split}
\end{equation}
where $\lambda$ serves as regularization strength.
Exemplarily, a resulting state trajectory is illustrated in Fig.~\ref{fig:xTraj_reg}.

\begin{figure}[htb!]
	\begin{center}
		\scalebox{0.5}{\input{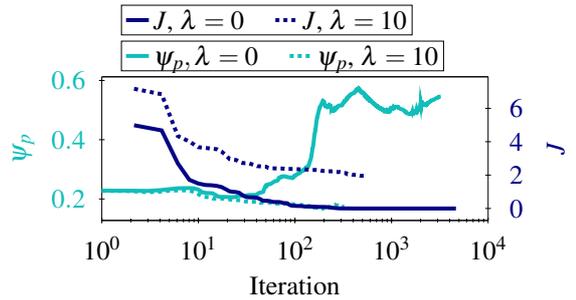}}
	\end{center}
	\caption{Comparison of regularization effect on $\psi_{p}$ and $J$. Both $\psi_{p}$ evolutions contain the initial space-filling at the first iteration with $\psi_p$=0.23.}
	\label{fig:SF_Evo_lambda10}
\end{figure}

\textbf{Remark.} The regularization technique leads to the stabilization of the local poles due to its relocation effect on data points in both LMs:
\begin{equation*} \label{eq:poles_reg}
	\begin{split}
		&\underset{i}{\max} \left| \mu_i(\underline{A}_{\text{LM,}1}) \right| = 0.78<1,\\
		&\underset{i}{\max} \left| \mu_i(\underline{A}_{\text{LM,}2}) \right| =0.89<1
	\end{split}
\end{equation*}\\
Next, we compare the evolution of loss $J$ and $\psi_{p}$ for the regularized and unregularized problem during the first split optimization, see Fig.~\ref{fig:SF_Evo_lambda10}.

As already mentioned, in the unregularized case, $\psi_{p}$ grows along the iterations, whereas the regularized curve shows a slight decrease. This means that the state trajectory \textit{expands} during optimization. Consequently, due to regularization, the loss cannot be reduced to~$J \approx 0$, because of the tradeoff between output error fit and space-filling of the distribution. At termination, it reaches a level of about~$J \approx 2$. Furthermore, the early-stopping effect caused by the penalty term is visible. The regularized problem is terminated after~324~iterations whereas the baseline runs till iteration~3232. This enables major savings in training time.

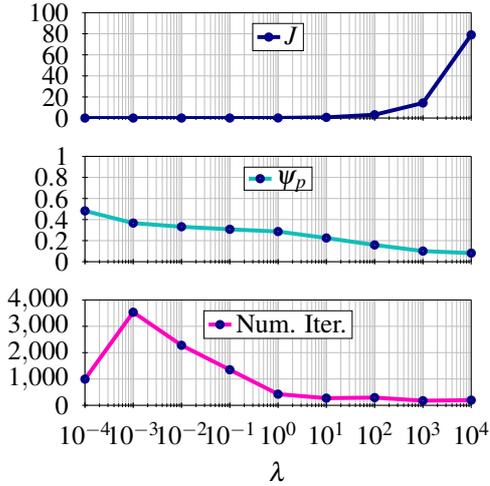
\begin{figure}[htb!]
	\begin{center}
		\scalebox{0.5}{\input{lossPsipIter-lambda_subplots_penalty2.tex}}
	\end{center}
	\caption{Loss, space-filling and iterations after optimization of problem~\eqref{eq:regopt} for various regularization strengths~$\lambda$.}
	\label{fig:lossPsip-lambda}
\end{figure}

For the choice of the hyperparameter $\lambda$, a grid search for values between $10^{-4}$ and $10^{4}$ is carried out. Figure~\ref{fig:lossPsip-lambda} visualizes $J$, $\psi_{p}$ and the number of iterations. 
The figure shows an increase of $J$ for stronger regularization, whereas $\psi_{p}$ can only be changed moderately due to the dynamic relations between data points. Since the initial state trajectory (iteration = 0, see light blue curve in Fig.~\ref{fig:xTraj}) shows suitable space-filling properties, it is sensible to choose $\lambda$ such that the final state trajectory yields a similar value for $\psi_{p}$.
As Fig.~\ref{fig:SF_Evo_lambda10} shows, the initial value for $\psi_p = 0.23$ is obtained for a linear model in balanced realization. This amount of space-filling can be roughly matched at convergence after the first split for the choice of regularization strength of $\lambda = 8$. Independent of the space-filling properties, $\lambda > 1$ are recommended to realize maximum advantage w.r.t.\ computational demand from the early-stopping (see Fig.~\ref{fig:lossPsip-lambda}, bottom diagram). 


\subsection{Desired Amount of Space-Filling}\label{sec:InitDeviation}
The proposed strategy requires a hyperparameter study to find a good value for~$\lambda$. If the desired value for space-filling is known a priori, e.g., keeping~$\psi_p$ from the initial model, an alternative approach is to penalize the \textit{deviation} between the current~$\psi_p$ and the desired one (named~$\psi_{p,\text{Target}}$). This regularized nonlinear problem can be stated as follows,
\begin{equation} \label{eq:nlp_initdev}
	\begin{split}
		\underset{\underline{\theta}}{\text{min}}\  &J(\underline{\theta}) + \lambda \cdot \Bigl(\psi_p(\underline{\theta}) - \psi_{p,\text{Target}}\Bigr)^2\\
		&\text{s.t. Dynamics in \eqref{eq:nlp}}.
	\end{split}
\end{equation}

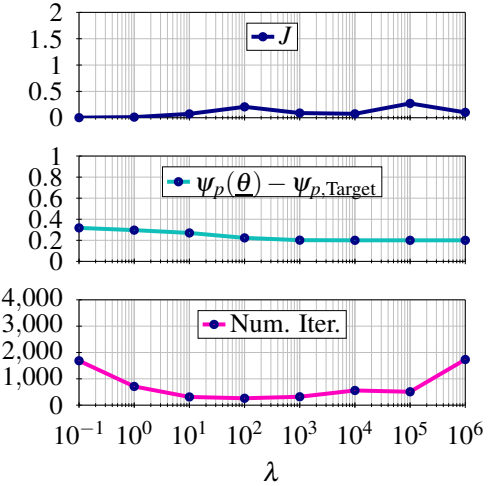
\begin{figure}[htb!]
	\begin{center}
		\scalebox{0.5}{\input{lossPsipIter-lambda_subplots2.tex}}
	\end{center}
	\caption{Optimization results of problem~\eqref{eq:nlp_initdev} for various values of~$\lambda$.}
	\label{fig:lossDeltaPsip-lambda}
\end{figure}

Here, a large value is chosen for the regularization strength $\lambda$, since it simply determines the tightness for matching~$\psi_{p,\text{Target}}$. For the demo process $\psi_{p,\text{Target}}$=0.23 is chosen, as it reflects the space-filling of the initial model. Figure~\ref{fig:lossDeltaPsip-lambda} shows the results of the study on~$\lambda$ for problem~\eqref{eq:nlp_initdev}. The loss~$J$ and the space-filling quality~$\psi_{p}$ are nearly independent of the choice of~$\lambda$. Therefore, it can be determined according to the other criteria. Consequently, it is chosen as $10^3$, being a tradeoff between initial space-filling and training time, measured by the number of iterations.


\section{\uppercase{Benchmark Dataset}}\label{sec:CaseStudy}
Next, we investigate the proposed regularization strategy on a well-known system identification benchmark, namely the Bouc-Wen Hysteretic System.
For details on the benchmark, we refer to~\citet{Schoukens2017}.

For brevity, we only show the results for the \textit{Desired Amount of Space-Filling} approach, see problem~\eqref{eq:nlp_initdev} in Section~\ref{sec:InitDeviation}. For the study, the space-filling goal is chosen as~$\psi_{p,\text{Target}}$=0.25. As for the demo process, the regularization strength is set to~$\lambda=10^3$.

\textbf{Remark.} The termination of LOLIMOT is based on the~\textit{validation} error. It is terminated when no significant improvements (25~\% of a normalized RMSE corresponding to an SNR of 40 dB) compared to the best loss in all previous iterations can be accomplished for three iterations. Then the simplest model within this error range is selected. 

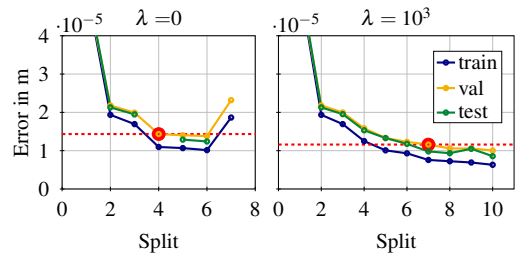
\begin{figure}[htb!]
	\begin{center}
		\scalebox{0.4}{\input{BWBench_RMSE-Splits_lambda0vs1000_subplots.tex}}
	\end{center}
	\caption{RMSE during LOLIMOT training. The \textit{best model} according to the validation error termination is marked in red.}
	\label{fig:BWBench_RMSE}
\end{figure}

In~Fig.~\ref{fig:BWBench_RMSE}, we compare the regularized and unregularized modeling cases regarding the RMSE of each split on training, validation, and test dataset.

According to the termination criterion, the best model for the unregularized case ($\lambda$=0) has four splits because the validation error increases due to the seventh split. This is caused by the splitting procedure of LOLIMOT (see Sec.~\ref{sec:LMSSN}). The addition of a LM, $n_{\text{LM},x} \rightarrow n_{\text{LM},x} + 1$, requires updating the centers and standard deviations of the Gaussians. Due to the normalization, the validity functions ${\Phi}^{[x]}_{j}$ are modified, which in turn influences the state trajectory and thus the output error. A poor initialization of the consecutive optimization of the split results and cannot be compensated. Consequently, the training error of the seven-split model is higher than the six-split model.

When evaluating the unregularized models on test data, the models obtained by split numbers four and seven show unstable behavior. In those cases, no sensible RMSE can be obtained, which is why their values are omitted from the figure. Within the regularized model ensemble, no instability could be detected regarding test data. Consequently, the robustness of LOLIMOT is improved.

Furthermore, it can be observed that regularization allows LOLIMOT to perform seven splits for the best model, compared to only four splits in the unregularized case. Regularization enables the algorithm to process the training dataset information with more LMs. Evaluation on the benchmark's test dataset yields an RMSE of~$0.981\cdot10^{-5}$ m for the regularized best model, whereas the unregularized one became unstable. In contrast to the stable splits, it can be stated that space-filling regularization enables better models. In comparison to other approaches, the achieved result is among the best for this benchmark~\citep{NBWG2025}.

\section{\uppercase{Conclusions}}\label{sec:Conclusion}
In this paper, we investigated the effect of nonlinear optimization on the extended input/state space data point distribution of nonlinear state space models. Two regularization approaches for penalizing poor space-filling quality are derived and tested, resulting in models with more meaningful and accurate local models. This is achieved by more local estimation and less interpolation as well as stable local behavior. Furthermore, higher modeling performance could be accomplished and the number of iterations could be reduced per optimization. Future lines of research will focus on the stable extrapolation behavior due to space-filling enforcement in the local model state space network.

\bibliographystyle{apalike}

\input{Example.bbl}

\end{document}

%% file: lossPsipIter-lambda_subplots_penalty2.tex
%
%
\definecolor{mycolor1}{rgb}{0.01176,0.01176,0.52157}%
\begin{tikzpicture}

\begin{axis}[%
width=4in,
height=1.1in,
at={(1.083in,6in)},
scale only axis,
separate axis lines,
xmode=log,
xmin=0.0001,
xmax=10000,
xminorticks=true,
xticklabels=\empty,
xticklabel style={yshift=-0.3cm},
xlabel style = {yshift = -1cm},
ymin=0,
ymax=100,
yticklabel style={xshift=-0.3cm},
ylabel style = {xshift = -1cm},
axis background/.style={fill=white},
xmajorgrids,
xminorgrids,
ymajorgrids,
legend style={draw=black, font=\fontsize{20}{10}\selectfont, at={(0.5,0.9)}, anchor=north},
label style={font=\fontsize{20}{10}\selectfont},
tick label style={font=\fontsize{20}{10}\selectfont}
]
\addplot [color=c209, line width=3.0pt, mark=o, mark options={solid, mycolor1}]
  table[row sep=crcr]{%
0.0001	3.32760181436242e-05\\
0.001	0.00013612840511712\\
0.01	0.0011390874601318\\
0.1	0.0100605389949209\\
1	0.0906035693812726\\
10	0.650992087717015\\
100	3.12787742803843\\
1000	14.3037494984105\\
10000	78.9660770388143\\
};
\addlegendentry{$J$}

\end{axis}

\begin{axis}[%
width=4in,
height=1.1in,
at={(1.083in,4.5in)},
scale only axis,
separate axis lines,
xmode=log,
xmin=0.0001,
xmax=10000,
xminorticks=true,
xticklabel style={yshift=-0.3cm},
xticklabels=\empty,
xlabel style = {yshift = -1cm},
ymin=0,
ymax=1,
yticklabel style={xshift=-0.3cm},
ylabel style = {xshift = -1cm},
axis background/.style={fill=white},
xmajorgrids,
xminorgrids,
ymajorgrids,
legend style={draw=black, font=\fontsize{20}{10}\selectfont, at={(0.5,0.9)}, anchor=north},
label style={font=\fontsize{20}{10}\selectfont},
tick label style={font=\fontsize{20}{10}\selectfont}
]
\addplot [color=c212, line width=3.0pt, mark=o, mark options={solid, mycolor1}]
  table[row sep=crcr]{%
0.0001	0.482459931489577\\
0.001	0.366728503178599\\
0.01	0.331750214896293\\
0.1	0.307651552501781\\
1	0.286401225023738\\
10	0.225159941639689\\
100	0.159192309083878\\
1000	0.101166946007122\\
10000	0.0814673672912027\\
};
\addlegendentry{$\psi_p$}

\end{axis}

\begin{axis}[%
width=4in,
height=1.1in,
at={(1.083in,3in)},
scale only axis,
separate axis lines,
xmode=log,
xmin=0.0001,
xmax=10000,
xminorticks=true,
xticklabel style={yshift=-0.3cm},
xlabel style = {yshift = -1cm},
xlabel={$\lambda$},
ymin=0,
ymax=4000,
yticklabel style={xshift=-0.3cm},
ylabel style = {xshift = -1cm},
axis background/.style={fill=white},
xmajorgrids,
xminorgrids,
ymajorgrids,
legend style={draw=black, font=\fontsize{20}{10}\selectfont, at={(0.5,0.9)}, anchor=north},
label style={font=\fontsize{20}{10}\selectfont},
tick label style={font=\fontsize{20}{10}\selectfont}
]
\addplot [color=c4, line width=3.0pt, mark=o, mark options={solid, mycolor1}]
  table[row sep=crcr]{%
0.0001	995\\
0.001	3534\\
0.01	2280\\
0.1	1351\\
1	424\\
10	272\\
100	293\\
1000	176\\
10000	196\\
};
\addlegendentry{Num. Iter.}

\end{axis}

\end{tikzpicture}%

%% file: lossPsipIter-lambda_subplots2.tex
%
%
\definecolor{mycolor1}{rgb}{0.01176,0.01176,0.52157}%
\begin{tikzpicture}

\begin{axis}[%
width=4in,
height=1.1in,
at={(1.083in,6in)},
scale only axis,
separate axis lines,
xmode=log,
xmin=0.1,
xmax=1000000,
xminorticks=true,
xticklabels=\empty,
xticklabel style={yshift=-0.3cm},
xlabel style = {yshift = -1cm},
ymin=0,
ymax=2,
yticklabel style={xshift=-0.3cm},
ylabel style = {xshift = -1cm},
axis background/.style={fill=white},
xmajorgrids,
xminorgrids,
ymajorgrids,
legend style={draw=black, font=\fontsize{20}{10}\selectfont, at={(0.5,0.9)}, anchor=north},
label style={font=\fontsize{20}{10}\selectfont},
tick label style={font=\fontsize{20}{10}\selectfont}
]
\addplot [color=c209, line width=3.0pt, mark=o, mark options={solid, mycolor1}]
  table[row sep=crcr]{%
0.1	0.00157893514167462\\
1	0.0114532495039841\\
10	0.0730885401743997\\
100	0.206962531235053\\
1000	0.0888563158428433\\
10000	0.0741210024470629\\
100000	0.271557693201884\\
1000000	0.101326745212633\\
};
\addlegendentry{$J$}

\end{axis}

\begin{axis}[%
width=4in,
height=1.1in,
at={(1.083in,4.5in)},
scale only axis,
separate axis lines,
xmode=log,
xmin=0.1,
xmax=1000000,
xminorticks=true,
xticklabels=\empty,
xticklabel style={yshift=-0.3cm},
xlabel style = {yshift = -1cm},
ymin =0,
ymax=1,
yticklabel style={xshift=-0.3cm},
ylabel style = {xshift = -1cm},
axis background/.style={fill=white},
xmajorgrids,
xminorgrids,
ymajorgrids,
legend style={draw=black, font=\fontsize{20}{10}\selectfont, at={(0.5,0.9)}, anchor=north},
label style={font=\fontsize{20}{10}\selectfont},
tick label style={font=\fontsize{20}{10}\selectfont}
]
\addplot [color=c212, line width=3.0pt, mark=o, mark options={solid, mycolor1}]
  table[row sep=crcr]{%
0.1	0.317989508173126\\
1	0.296697451705243\\
10	0.269958153590022\\
100	0.223155446350655\\
1000	0.20141840590717\\
10000	0.200121378618548\\
100000	0.200035101033306\\
1000000	0.199995360517098\\
};
\addlegendentry{$\psi_p(\underline{\theta}) - \psi_{p,\text{Target}}$}

\end{axis}

\begin{axis}[%
width=4in,
height=1.1in,
at={(1.083in,3in)},
scale only axis,
separate axis lines,
xmode=log,
xmin=0.1,
xmax=1000000,
xminorticks=true,
xticklabel style={yshift=-0.3cm},
xlabel style = {yshift = -1cm},
xlabel={$\lambda$},
ymin=0,
ymax=4000,
yticklabel style={xshift=-0.3cm},
ylabel style = {xshift = -1cm},
axis background/.style={fill=white},
xmajorgrids,
xminorgrids,
ymajorgrids,
legend style={draw=black, font=\fontsize{20}{10}\selectfont, at={(0.5,0.9)}, anchor=north},
label style={font=\fontsize{20}{10}\selectfont},
tick label style={font=\fontsize{20}{10}\selectfont}
]
\addplot [color=c4, line width=3.0pt, mark=o, mark options={solid, mycolor1}]
  table[row sep=crcr]{%
0.1	1687\\
1	711\\
10	312\\
100	262\\
1000	319\\
10000	558\\
100000	508\\
1000000	1734\\
};
\addlegendentry{Num. Iter.}

\end{axis}

\end{tikzpicture}%

%% file: BWBench_RMSE-Splits_lambda0vs1000_subplots.tex
%
%
\definecolor{mycolor1}{rgb}{0.01176,0.01176,0.52157}%
\definecolor{mycolor2}{rgb}{0.97255,0.69020,0.00000}%
\definecolor{mycolor3}{rgb}{0.00000,0.52549,0.20784}%
\begin{tikzpicture}

\begin{axis}[%
width=1.22*2.5in,
height=2in,
at={(2.8in,0.458in)},
scale only axis,
xmin=0,
xmax=11,
xlabel={Split},
xticklabel style={yshift=-0.3cm},
xlabel style = {yshift = -1cm},
ymin=0,
ymax=0.00004,
yticklabel style={xshift=-0.3cm},
yticklabel={\empty},
axis background/.style={fill=white},
title style={font=\fontsize{20}{10}\selectfont,yshift = 0cm},
title={$\lambda=10^3$},
xmajorgrids,
ymajorgrids,
label style={font=\fontsize{20}{10}\selectfont},
tick label style={font=\fontsize{20}{10}\selectfont}
]
\addplot [color=mycolor1, line width=2.0pt, mark=o, mark options={solid, mycolor1}]
  table[row sep=crcr]{%
0	0.000152058782999731\\
1	5.29826012449774e-05\\
2	1.93518048057228e-05\\
3	1.69396355786944e-05\\
4	1.25007803291055e-05\\
5	1.00868286221356e-05\\
6	9.30568560342238e-06\\
7	7.5773037266913e-06\\
8	7.24481444436265e-06\\
9	6.91483909420754e-06\\
10	6.31196927614042e-06\\
};

\addplot[
only marks,
color=red,
mark=*,
mark options={scale=3.0},
]
coordinates {(7,1.15978445052178e-05)};

\addplot[
color=red,
line width=2pt,
domain=0:11,
forget plot,
dashed
] coordinates {(0,1.15978445052178e-05) (11,1.15978445052178e-05)};

\addplot [color=mycolor2, line width=2.0pt, mark=o, mark options={solid, mycolor2}]
  table[row sep=crcr]{%
0	0.000144737787652005\\
1	5.30524444375402e-05\\
2	2.1797175801197e-05\\
3	1.99728234608898e-05\\
4	1.58298160890524e-05\\
5	1.33247467744013e-05\\
6	1.22960848170095e-05\\
7	1.15978445052178e-05\\
8	1.05698851342595e-05\\
9	1.05782571342417e-05\\
10	1.0050677760276e-05\\
};

\addplot [color=mycolor3, line width=2.0pt, mark=o, mark options={solid, mycolor3}]
  table[row sep=crcr]{%
0	0.000148146660235327\\
1	5.44627919545349e-05\\
2	2.12927534537499e-05\\
3	1.95278091741358e-05\\
4	1.53236124358358e-05\\
5	1.32845705742568e-05\\
6	1.17831937735957e-05\\
7	9.80574640685461e-06\\
8	9.3430461891586e-06\\
9	1.04965687199865e-05\\
10	8.56562008166861e-06\\
};


\end{axis}

\begin{axis}[%
width=2.5in,
height=2in,
at={(0in,0.458in)},
scale only axis,
xmin=0,
xmax=8,
xlabel={Split},
xticklabel style={yshift=-0.3cm},
xlabel style = {yshift = -1cm},
ymin=0,
ymax=0.00004,
yticklabel style={xshift=-0.3cm},
ylabel={Error in m},
ylabel style = {yshift = 0.3cm},
axis background/.style={fill=white},
title style={font=\fontsize{20}{10}\selectfont,yshift = 0cm},
title={$\lambda=$0},
xmajorgrids,
ymajorgrids,
legend style={
	draw=black,
	font=\fontsize{20}{10}\selectfont,
	at={(2.1,0.9)},
	anchor=north,
	legend columns=1,      
	/tikz/column sep=0pt,  
	/tikz/row sep=5pt      
},
legend cell align={left},
label style={font=\fontsize{20}{10}\selectfont},
tick label style={font=\fontsize{20}{10}\selectfont}
]
\addplot [color=mycolor1, line width=2.0pt, mark=o, mark options={solid, mycolor1}]
  table[row sep=crcr]{%
0	0.000152058731803351\\
1	5.29826540507372e-05\\
2	1.93519105442519e-05\\
3	1.69385947534906e-05\\
4	1.09684396631948e-05\\
5	1.0672987950322e-05\\
6	1.01439754409682e-05\\
7	1.86466159286525e-05\\
};
\addlegendentry{train}

\addplot[
only marks,
color=red,
mark=*,
mark options={scale=3.0},
forget plot
] coordinates {(4,1.43389752729206e-05)};

\addplot[
color=red,
line width=2pt,
domain=0:11,
forget plot,
dashed
] coordinates {(0,1.43389752729206e-05) (11,1.43389752729206e-05)};

\addplot [color=mycolor2, line width=2.0pt, mark=o, mark options={solid, mycolor2}]
  table[row sep=crcr]{%
0	0.000144735472775714\\
1	5.30537451668422e-05\\
2	2.18010543315827e-05\\
3	1.99163478020736e-05\\
4	1.43389752729206e-05\\
5	1.40244199640094e-05\\
6	1.38058137810759e-05\\
7	2.31799234260421e-05\\
};
\addlegendentry{val}

\addplot [color=mycolor3, line width=2.0pt, mark=o, mark options={solid, mycolor3}]
  table[row sep=crcr]{%
5	1.28812532178557e-05\\
6	1.24152168901048e-05\\
};
\addplot [color=mycolor3, line width=2.0pt, mark=o, mark options={solid, mycolor3}]
table[row sep=crcr]{%
	0	0.000148146542858996\\
	1	5.44624814500239e-05\\
	2	2.13004484557189e-05\\
	3	1.94690422530286e-05\\
};
\addlegendentry{test}

\end{axis}

\end{tikzpicture}%